\documentclass[preprint,12pt]{elsarticle}
\usepackage{times}
\usepackage{soul}
\usepackage{url}
\usepackage[hidelinks]{hyperref}
\usepackage[utf8]{inputenc}
\usepackage[small]{caption}
\usepackage{graphicx}
\usepackage{amsmath}
\usepackage{amsthm}
\usepackage{booktabs}
\usepackage{algorithm}
\usepackage{algorithmic}
\usepackage{subfigure}
\usepackage{amsmath}
\usepackage{amssymb}
\usepackage{bm}
\usepackage{lineno}
\usepackage{color}

\journal{Pattern recognition}
\begin{document}

\begin{frontmatter}

\title{Reasoning Structural Relation for Occlusion-Robust Facial Landmark Localization}

\author[a]{Congcong Zhu}
\ead{congcongzhu@shu.edu.cn}
\author[a]{Xiaoqiang Li\corref{cor1}}
\cortext[cor1]{Corresponding author is Xiaoqiang Li.}
\ead{xqli@shu.edu.cn}
\author[a]{Jide Li}
\ead{iavtvai@shu.edu.cn}
\author[a]{Songmin Dai}
\ead{laodar@shu.edu.cn}
\author[a]{Weiqin Tong}
\ead{wqtong@shu.edu.cn}

\address[a]{School of Computer Engineering and Science, Shanghai University, Shanghai, 200444, China}

\begin{abstract}
In facial landmark localization tasks, various occlusions heavily degrade the localization accuracy due to the partial observability of facial features. This paper proposes a structural relation network (SRN) for occlusion-robust landmark localization. Unlike most existing methods that simply exploit the shape constraint, the proposed SRN aims to capture the structural relations among different facial components. These relations can be considered a more powerful shape constraint against occlusion. To achieve this, a hierarchical structural relation module (HSRM) is designed to hierarchically reason the structural relations that represent both long- and short-distance spatial dependencies. Compared with existing network architectures,the HSRM can efficiently model the spatial relations by leveraging its geometry-aware network architecture, which reduces the semantic ambiguity caused by occlusion. Moreover, the SRN augments the training data by synthesizing occluded faces. To further extend our SRN for occluded video data, we formulate the occluded face synthesis as a Markov decision process (MDP). Specifically, it plans the movement of the dynamic occlusion based on an accumulated reward associated with the performance degradation of the pre-trained SRN. This procedure augments hard samples for robust facial landmark tracking. Extensive experimental results indicate that the proposed method achieves outstanding performance on occluded and masked faces. Code is available at \url{https://github.com/zhuccly/SRN}
\end{abstract} 

\end{frontmatter}


\section{Introduction}
\label{}
Facial landmark localization, including image-based detection and video-based tracking, aims to localize facial keypoints for a given image or video clip. It plays an important role in many facial analyses, such as 3D face reconstruction~\cite{jiang20183d,gecer2019ganfit,PRNET}, face recognition~\cite{zhao2018towards,kang2018pairwise} and face editing~\cite{song2019geometry,wang2019example}. In the last decade, many studies based on deep learning \cite{MDM,SBR,3DDFA,FAN,LAB} have achieved great success in this field. Despite this, various occlusions still heavily degrade localization performance. There are two leading causes. First, most methods \cite{SDM,LBF,SBR,3DDFA,FAN,MDM,CFSS,LAB} rely completely on appearance information that may be perturbed by various occlusions; Second, general neural network architectures, such as convolutional neural networks (CNNs) and multi-layer perceptrons (MLPs), have difficulties reasoning the structural relations of facial landmarks~\cite{santoro2017simple}. As a result, structural abnormalities of landmarks are observed when these methods are applied to hard samples containing severe occlusions. To address these problems, some studies \cite{MDM,liu2019semantic} impose a shape constraint by jointly predicting all landmarks, but this rough shape constraint may cause the entire face shape to deviate from the ground-truth landmarks. Other methods such as LAB~\cite{LAB} and PCD-CNN~\cite{kumar2018disentangling} introduce the geometric prior, which is predicted by additional branch networks, making the models robust to occlusion. However, they predict the geometric information without learning the mutual relations among landmarks, which may accumulate incorrect geometric information. RCPR \cite{RCPR} provides a small face dataset (COFW) focused on occlusion, containing $1345$ face images with $29$ sparse landmarks. However, any performance evaluation using COFW is limited because of insufficient data. Furthermore, the annotation with $29$ landmarks cannot be merged into other datasets that use $68$ landmarks for dense landmark localization tasks. Therefore, handling diverse occluded faces in unconstrained environments remains a challenge. 

In this study, we propose a structural relation network (SRN) that reasons the structural relations among different facial components~(\textit{e.g.}, nose, eyes and brows). Structural relations can be used to build long- and short-distance spatial dependencies that enhance robustness to various occlusions. Santoro \textit{et al.}~\cite{santoro2017simple} demonstrated that seemingly simple relational inferences are remarkably difficult to learn even for powerful neural network architectures such as CNNs and MLPs. Therefore, our SRN employs novel network architectures with a hierarchical structure whose computations focus explicitly on relational reasoning. To achieve this, we designed a hierarchical structural relation module (HSRM) that hierarchically divides the facial components into different relational groups by following the inherent structure of the entire face, as shown in Figure \ref{Idea}. Moreover, we diversify occluded faces by applying a noise patch surrounding a randomly selected ground-truth landmark. For dynamic faces that lack examples of occluded faces, we hope to augment hard samples with occlusion to significantly reinforce the robustness of our SRN to achieve occlusion-robust tracking. However, there are no supervision signals available in existing datasets that can be used to augment hard samples. Therefore, we formulate the occluded face synthesis as a Markov decision process (MDP), which determines the trajectory of dynamic occlusions in sequential faces. Specifically, we synthesize candidate occluded faces and use deep Q-learning (DQL)~\cite{mnih2013playing} to select the "hardest" samples from these candidate faces. This procedure synthesizes the hard occluded faces that maximize the performance degradation of the pre-trained SRN. The SRN is then fine-tuned to handle these occluded faces. We summarize our main contributions as follows: 
\begin{figure*}[t]
	\begin{center}
		\includegraphics[width=1\linewidth]{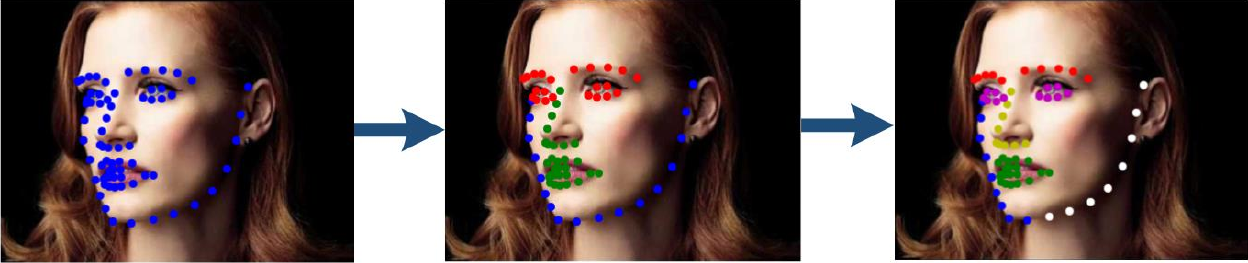}
	\end{center}
	\caption{Insight regarding the hierarchical dependency relation. We hierarchically divide the entire face into six semantic components. In each image, landmarks of the same color indicate that they were divided into the same component. The proposed SRN captures the hierarchical structure relations among facial components.}
	\label{Idea}
\end{figure*}
\begin{itemize}
\item We propose a structural relation network (SRN) for occlusion-robust facial landmark localization. It exploits the structural relations among different facial components as a more powerful shape constraint and synthesizes occluded faces for data augmentation.
\item We design a hierarchical structural relation module (HSRM) to implement the structured network architecture, which reasons the structural relations to build both long- and short-distance spatial dependencies.
\item To further extend our SRN to occluded video data, we formulate the occlusion synthesis of dynamic faces as a Markov decision process (MDP), which interacts with the pre-trained SRN via an accumulated reward to synthesize hard occluded videos that are used to fine-tune the SRN. 
\end{itemize} 

\section{Related Work}

Facial alignment is mainly performed in two application scenarios, image-based and video-based. Image-based methods~\cite{CootsAAM,SDM,MDM,3DDFA,CFSS,LAB,FAN} localize facial landmarks on a single image, with the aim of reducing spatial misalignments. In~\cite{ESR,PCR,zhang2014coarse,CFSS}, facial landmark localization was implemented using a cascaded regression process, which refines the initial shape to the final shape in a coarse-to-fine manner. After applying CNNs, studies such as~\cite{HongMPT13,TSR,MDM} achieved competitive performance by extracting discriminative features from pixels and modeling the nonlinear relation between facial images and landmarks. With the large pose issues taken into consideration, 3D face information has been introduced to address large-pose issues, which fits a 3D morphable model (3DMM) to a 2D image~\cite{FAN,3DMM,3DDFA,PIFA,JOINT}. However, methods that apply 3DMM are computationally intensive. To address the occlusion problem, RCPR~\cite{RCPR} reduces exposure to outliers by detecting occlusions explicitly to construct robust shape-indexed features. PCD-CNN~\cite{kumar2018disentangling} was developed as a cascade local prediction method using a Dendritic CNN, which ignores the mutual constraints between different components of the face. 3DDE~\cite{VALLE2019102846} employs a stacked deep U-net to predict coarse probability maps of landmarks, and then it fits a 3D face model to the maps. Finally, the regressor implicitly imposes a prior face shape to refine the facial shape using a coarse-to-fine cascaded regression. Both deep U-net and 3D face models are computationally expensive. The following methods handle occlusion faces by stacking additional parameters and subnetworks. LUVLi~\cite{kumar2020luvli} jointly predicts landmark locations, associated uncertainties among these predicted locations, and landmark visibility. It models these as mixed random variables and estimates them using a four-stage stacked hourglass network (HG). Look at boundary (LAB)~\cite{LAB} imposes global geometric constraints among all landmarks by introducing boundary information, in which estimating the boundary heatmap is computationally complex. PropNet~\cite{huang2020propagationnet} stacks massive parameters to propagate landmark heatmaps to boundary heatmaps. Comparing the HG based network to LAB~\cite{LAB} which also considers edge information, its parameters are approximately four times those of LAB~\cite{LAB}.

Facial landmark tracking, \textit{a.k.a.}, video-based facial landmark localization, aims to stably predict sequential facial landmarks for a given video clip~\cite{DBLP:conf/eccv/Sanchez-LozanoM16,DBLP:conf/eccv/GuoLZ18,DBLP:journals/pami/LiuLFZ18,STKI,FHR}. Some works~\cite{DBLP:journals/pami/LiuLFZ18,DBLP:conf/eccv/PengFWM16} studies have used networks to memorize temporal information across frames. However, they may accumulate occlusion information that leads to landmark jitter. To address this issue, SBR~\cite{SBR} introduced a cycle-consistency check for self-supervised learning, which establishes the dense correspondences of landmarks between frames. Moreover, the employed tracker forward localizes facial landmarks and then backward evaluates the predictions to augment the training loss function by constraining tracking to a cycle-consistent process. STRRN~\cite{zhu2020towards} disentangles facial geometry relations for every static frame and simultaneously enforces bi-directional cycle-consistency across adjacent frames, thus modeling the intrinsic spatial-temporal relations of raw face sequences. In this method, the detector and tracker are trained using interactive learning to enhance tracking consistency. FHR ~\cite{FHR} uses a stabilization algorithm based on a probability model to eliminate jitter, but lacks the capability to model large movements in short videos. This is because FHR does not consider inter-frame temporal dependency, so it has difficulties addressing the problem of heavy occlusions in motion. FAB~\cite{sun2019fab} aims to handle motion-blurred videos by utilizing eight residual blocks to build an hourglass network for predicting boundary maps. Additionally, two convolutional layers and four residual blocks are used to generate a de-blurred sharp image, which form a pre-activated Resnet-18 as FAB's replaceable facial landmark detection network for landmark detection. The complex networks and a large number of parameters make it difficult to achieve a satisfactory tracking speed. The tracker used by~\cite{yin2020exploiting} also uses a four-stage stacked HG and adds an additional neural network as a landmark regressor. The input to this stacked model based tracker is a multi-frame sequence, which leads to a fatal weakness in tracking speed. Although manual occlusions are inserted in this method, the model is not specifically optimized for occlusions, nor is it evaluated on any occlusion data.  

As a result, current methods~\cite{RCPR,VALLE2019102846,kumar2020luvli,LAB,huang2020propagationnet} to address occlusions are mainly developed using image-based face alignment, and cannot be easily extended to achieve stable tracking involving occluded videos. Therefore, there is an urgent need to develop a lightweight and easily extendable robust facial landmark localization method for occlusions.

\section{Approach}
\subsection{Structural Relation Networks}
\label{section3.1}
\begin{figure*}[h]
	\begin{center}

		\includegraphics[width=0.95\linewidth]{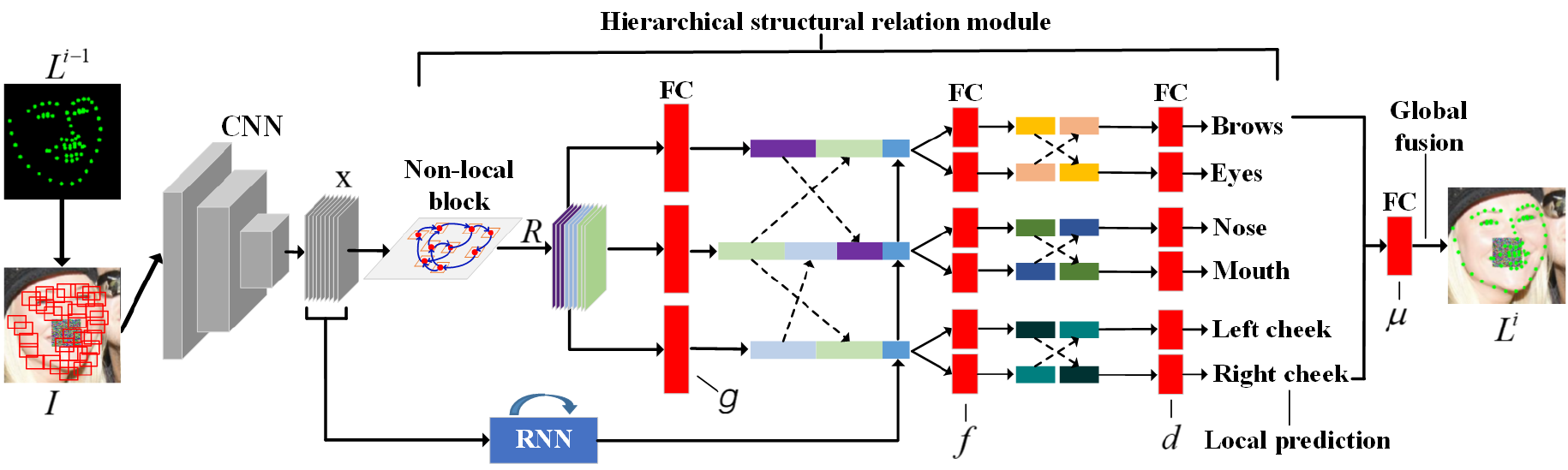}
	\end{center}
	\caption{Illustration of the proposed SRN. Given a facial image and corresponding ground-truth landmarks, our SRN first inserts a noise occlusion patch surrounding a ground-truth landmark selected randomly and then predicts facial landmarks by refining the initial landmarks using cascaded regression. Taking the $i$-th regression iteration as an example, our SRN hierarchically reasons the structural relations by employing the hierarchical structural relation module (HSRM). In addition, $g$, $f$, $d$ and $\mu$ represent different hierarchies conditioned on global appearance information. Recurrent neural networks (RNNs) memorize the global appearance information to be embedded into each facial component group.}
	\label{net_v}
\end{figure*}

Facial landmarks are facial key points that indirectly contain the biological information and describe the physiological structure of the human face. Exploiting the inherent structural relations of the entire face can efficiently improve a model's robustness against various occlusions. Based on this fact, we propose a structural relation network (SRN) to reason structural relations among facial components, as shown in Figure~\ref{net_v}.	Next, we provide the details regarding the proposed SRN.

Given a face image $I$, we patch a noise occlusion around a randomly selected ground-truth landmark. Our SRN performs cascaded regression based on the facial patch set, which includes 68 patches corresponding to 68 landmarks in fixed index order. We let $L=[l_{1},l_{2},\dots,l_{N}]$ $\in \mathbb{R}^{2\times N}$ represent a face shape vector with $N$ facial landmarks, where $l_n$ denotes the coordinates of the $n$-th landmark. Let $L^{i}$ denote the predicted result for the $i$-th iteration, which is obtained based on the predicted result of the previous iteration
\begin{eqnarray} 
\label{regession}
L^{i} = L^{i-1} + \bigtriangleup L^{i},
\end{eqnarray}     
where $\bigtriangleup L^{i}$ represents the shape residual at the $i$-th iteration. The maximum number of iterations is set to $3$ in our task. The shape residual is predicted based on local patches surrounding the landmarks of the previous results $L^{i-1}$. These patches, extracted from the occluded face image using a shape-indexed approach~\cite{MDM}, are fed into the proposed SRN . $\bigtriangleup L^{i}$ can be expressed as    
\begin{eqnarray} 
\label{prediction}
\bigtriangleup L^{i} = \varphi (\rho  (F | L^{i-1})),
\end{eqnarray} 

where $\rho (\cdot)$ is a patch-extracting operation that crops facial patches, and $\varphi (\cdot)$ is the prediction operation of the SRN. For the first iteration, the initial landmarks are represented by the mean shape of all the training samples.

To capture long- and short-distance spatial dependencies, the hierarchical structural relation module (HSRM) is designed to hierarchically reason the structural relations, as illustrated in Figure~\ref{net_v}. Specifically, facial patches are fed into three convolutional layers to output the feature map set $\mathbf{x}$ with $k*N$ maps, where $k$ is the number of kernels in the final convolutional layer and $N$ denotes the number of facial patches. Because it is difficult for a CNN with a fixed receptive field to reason the structural relations, we use a non-local operation \cite{buades2005non} to capture non-local structural relations directly by computing interactions between any feature map pair in $\mathbf{x}$
\begin{eqnarray} 
\label{non-local}
\mathbf{y}_{u}=\frac{1}{\mathcal{C}(\mathbf{x})} \sum_{\forall v} \sigma\left(\mathbf{x}_{u}, \mathbf{x}_{v}\right) \pi\left(\mathbf{x}_{v}\right).
\end{eqnarray} 
Here $\mathbf{y}$ is the output signal of the same size as $\mathbf{x}$. $u$ denotes the index of an output position in the feature map set $\mathbf{x}$ (whose response is to be computed), and $v$ is the index that enumerates all possible positions.  $\sigma(\cdot,\cdot)$ is a function of the feature embeddings for all pairs. It calculates the dot-product similarity, representing a relation. The unary function $\pi$ computes a representation of the input signal at position $v$. The response is normalized by a factor ${\mathcal{C}(\mathbf{x})}$, which represents the number of all positions in $\mathbf{x}$. For our method, we use a non-local block \cite{Wang_2018_CVPR} to achieve the non-local operation. Then, we obtain the global relation feature map set $R$ by calculating the residual connection  
\begin{eqnarray} 
\label{non-local block}
R = \mathcal{F}(\mathbf{y}) + \mathbf{x},
\end{eqnarray} 
 
where $\mathcal{F}(\cdot)$ denotes the $1*1$ convolutions. The residual connection allows the non-local block to be considered as an increment operation that can be inserted into any pre-trained model without disrupting its initial objective.

To further reason the neighborhood relations involved in detailed facial landmark localization, the HSRM captures hierarchical spatial dependencies. It disentangles the non-local feature map set $R$ into three neighborhood maps based on the facial patch indexes: the ocular map $R_{o}$ including eye ($12$ landmarks) and brow ($10$ landmarks) groups, the snout map $R_{s}$ including nose ($9$ landmarks) and mouth ($20$ landmarks) groups, and the cheek map $R_{c}$ including left ($9$ landmarks) and right ($8$ landmarks) cheek groups. Hence, we can reorganize the feature map set $R$ into three neighborhood feature maps and six component groups by leveraging this structured network architecture, which conforms to the physiological structure of the entire face, as shown in Figure~\ref{net_v}. Thus, geometrical relations are captured within the hierarchical structure. The SRN considers relations across all hierarchies (conditioned on global appearance information) and integrates all of these relations to constrain the face shape. Moreover, recurrent neural networks (RNNs) are introduced to memorize the global appearance information at each iteration step and embed this information into each group. For the RNN layer, we formulate the hidden state $h^{i}$ for iteration step $i$ as 
\begin{eqnarray} \label{RNN}
h^{i}=\mathrm{tanh}((\mathbf{x} \oplus h^{i-1})*W_{1}+b_{1}),
\end{eqnarray}
where $\mathrm{tanh}$ is the activation function, $W_{1}$ denotes the weight matrix of the input-to-hidden fully connected layer, $b_{1}$ is bias and $\oplus$ is concatenation operation. It is helpful to acquire more context as reference information. Taking the eye group as an exmple, the operation of the HSRM can be formulated as 
\begin{eqnarray} 
 \bigtriangleup L_{e}=d_{ e}\left ( f_{ e}\left ( g_{o}(R)\oplus g_{ s}(R) \oplus h \right )\oplus f_{b}\left ( g_{o}(R)\oplus g_{ s}(R) \oplus h \right )\right ), 
\end{eqnarray}  
where $g_{o}(\cdot)$ and $g_{s}(\cdot)$ disentangle the non-local features into ocular and snout feature maps, and $f_{e}(\cdot)$ and $f_{b}(\cdot)$ subsequently disentangle the ocular features into the group features representing eyes and brows. Here, the index $i$ was ignored for simplicity. 
 
Finally, we merge all landmarks into a global shape vector using a concatenation operation. We define the global shape integration as
\begin{eqnarray} 
\bigtriangleup L = \mu \left(\bigcup_{m=1}^{M}\bigtriangleup L_{m}\right),
\end{eqnarray}
where $M$ denotes the number of facial groups, $\mu(\cdot)$ imposes a point-to-point constraint by jointly regressing all landmarks, thus fine-tuning the global shape vector, and $\bigcup$ represents the merging of all landmarks into a face-shape vector.

Mathematically, our SRN optimizes the following objective function:
\begin{eqnarray}
\label{8}
\text{min}\sum_{i=1}^{I}\left \| \bar L-\left ( L^{i-1}+\bigtriangleup L^{i} \right ) \right \|_{2}^{2},
\end{eqnarray}
where $I$, $ \bar L$, $L^{i-1}$, and $ \bigtriangleup L^{i} $ denote the maximum iteration step, the ground-truth landmarks, the results of the previous iteration, and the facial shape residual at the $i$-th iteration, respectively. The minimization is computed over all parameters of our SRN. 
\subsection{Occlusion-Robust Facial Landmark Tracking}
\label{section3.2}
\begin{figure*}[h]
	\begin{center}
		\includegraphics[width=1\linewidth]{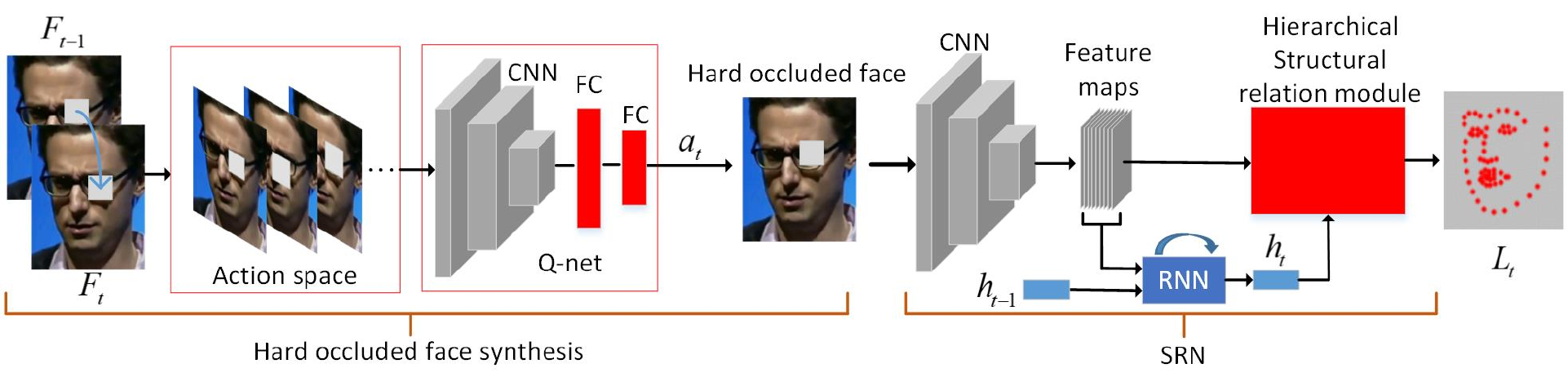}
	\end{center}
	\caption{Illustration of the proposed occlusion-robust facial landmark tracking scheme. Q-net denotes the action value function that evaluates each occluded face candidate. For landmark tracking, the RNN module memorizes temporal information flowing across frames. This RNN is different from the previous RNN introduced in Section~\ref{section3.1}, which memorizes spatial information.}
	\label{Nets_video}
\end{figure*}

Our occlusion-robust facial landmark tracking method involves two main stages: (1) hard occluded face synthesis (HOFS) for data augmentation, and (2) fine-tuning the pre-trained SRN using synthesized occluded faces (see Figure \ref{Nets_video}). The former aims to synthesize hard samples during training. However, there are no supervision signals available in existing datasets for synthesizing hard samples. To address this, we formulate the occluded face synthesis as a Markov decision process (MDP) in dynamic faces, which employs the action value function as the proxy for occluded face evaluation. Given a frame $F_{t}$ at time $t$ and the ground-truth landmarks $\bar L_{t}$, we produce an occlusion around three adjacent landmarks: $index-1$, $index$ and $index+1$. These three occluded faces are fed into a deep Q-net, designed by applying the deep Q-learning \cite{mnih2013playing, mnih2015human}, to select the hard sample that maximizes the performance degradation of the pre-trained SRN. Applying reinforcement learning directly using consecutive faces is inefficient due to two reasons: 1) strong correlations between adjacent frames in the same video clip; and 2) lack of data diversity occurs because the "300 videos in the wild" (300VW) dataset contains only 64 training videos. To address this problem, we utilize the deep Q-net (DQN)  proposed by~\cite{mnih2013playing,mnih2015human} to develop our Q-net. This method provides a major advantage for video data, as its "experience replay" feature~\cite{lin1992reinforcement} can break correlations between adjacent frames, thus reducing the variance of the update.

\begin{figure*}[h]
	\begin{center}
		\includegraphics[width=1\linewidth]{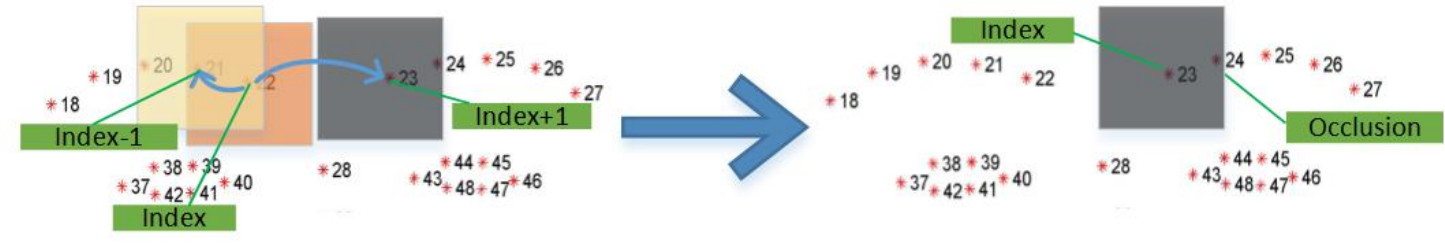}
	\end{center}
	\caption{The defined actions for the ocular region. The boxes represent candidate occlusions that aim to degrade the performance of the pre-trained SRN. The gray box represents the selected action of the current frame.}
	\label{face}
\end{figure*}

\textbf{Action}: We define an action as
\begin{eqnarray} 
\label{equ:reward}
a_{t}= \begin{cases} a_{t}^{0}, & \text {if the occlusion shifts to the $index-1$ landmark}, 
\nonumber \\ a_{t}^{1} , &\text {if the occlusion remains on the $index$ landmark}, 
\nonumber \\ a_{t}^{2} , &\text {if the occlusion shifts to the $index+1$ landmark}. 
\nonumber \end{cases}
\end{eqnarray}
The action space includes three candidate actions that represent occluded faces with a noise patch surrounding the $index-1$, $index$ or $index+1$ landmarks. Here, $index$ is provided by the selected action of the previous frame and is randomly specified in the first frame. These candidate actions are associated with whether the occlusion is shifts between adjacent facial components, as shown in Figure~\ref{face}. 

\textbf{State}:
$S_{t}^{p}$ represents facial patches observed on the occluded face synthesized by the current candidate action $a_{t}^{p}$, where $p \in [0,1,2]$ denotes one of the candidate actions. Subsequently, the action value function $Q(S_{t}^{p}, a_{t}^{p})$ computes the Q values corresponding to the state-action pair. In our method, Q-net denotes the action value function. Based on $S_{t}^{p}$, Q-net outputs an action value Q. We then select the optimal action corresponding to the highest Q value that is associated with the maximum performance degradation of the pre-trained SRN. By leveraging the ground-truth landmarks $\bar L_{t}$, the state $S_{t}^{p}$ observed at the $p$-th candidate action is formulated as
\begin{eqnarray} 
\label{state}
S_{t}^{p}=\rho(F_{t},a_{t}^{p}|\bar L_{t}),
\end{eqnarray}
where $\rho(\cdot)$ denotes the patch-extracting operation using a shape-indexed manner. 

\textbf{State Transition}:
The state is observed on the synthesized occluded face by shifting the occlusion among landmarks: $index-1$, $index$, and $index+1$. Therefore, the state of the $t+1$ frame is related to the index transition.
\begin{eqnarray} 
\label{equ:index}
index= \begin{cases} index-1, & \text {if } \quad a_{t}^{0}=\mathrm{arg \ max}( Q (S_{t}^{p},a_{t}^{p})), 
\nonumber \\ index , &\text {if } \quad a_{t}^{1}=\mathrm{arg \ max}( Q (S_{t}^{p},a_{t}^{p})),  
\nonumber \\ index+1 , &\text {otherwise },
\nonumber \end{cases}
\end{eqnarray} 
where $\mathrm{arg \ max}(\cdot)$ denotes the selection of the optimal action corresponding to the highest action value. The new state is observed based on the updated $index$.   

\textbf{Reward}: 
The reward $r_{t}$ reflects the performance degradation of the pre-trained SRN caused by the synthesized face. It is defined as the point-to-point Euclidean error~\cite{300W}:
\begin{eqnarray} 
\label{equ:e}
r_{t}=\frac{\sum_{n=1}^{N}\left\|L_{t}^{n}-\bar{L}_{t}^{n}\right\|^2_2}{N \cdot \eta},
\end{eqnarray}
where $N$, $\eta$, $L_{t}^{n}$ and $\bar{L}_{t}^{n}$ denote the number of landmarks, the normalizing factor, the predicted landmarks and the ground-truth landmarks, respectively. This reward encourages Q-net to explore the weaknesses of the pre-trained SRN.

\textbf{Reinforcement Learning Stage}:
Our Q-net evaluates all candidate occluded faces and outputs the corresponding action value. We optimize the parameters of Q-net by updating the action value function $Q(S,a)$. Omitting the subscript $p$ for simplicity, $Q(S,a)$ is iteratively updated according to the Bellman equation:
\begin{eqnarray}
Q\left (S_{t},a_{t}\right) = r_{t} + \gamma \text{max}Q\left (S_{t+1},a_{t+1}\right),
\end{eqnarray}
where $\gamma = 0.9$ and $\text{max}Q\left(S_{t+1},a_{t+1}\right) $ denote the discount factor and the future maximum benefit, respectively. 

For the Q-net, we minimize the following loss to update all parameters: 
\begin{eqnarray}
L_{Q} = \mathbb{E}\left [Q\left (S_{t},a_{t}\right)-\left ( r_{t} +\gamma \text{max}Q\left (S_{t+1},a_{t+1}\right) \right ) \right ]^{2}.
\end{eqnarray}

\textbf{Experience Replay}:
		Experience replay is an easy-to-implement training method. It averages the behavior distribution over many of its previous states and smooths out the learning process. The experience replay strategy used in the original DQN~\cite{mnih2013playing}, which stores the agent's last experience at each step (queue storage with a capacity limit of $N_{total}$) and samples uniformly at random from these experience when performing updates. Unlike~\cite{mnih2013playing}, we store $N_{clip}*K$ experiences of the Q-net on last $K=3$ video clips (multi-queue storage) to diversify experiences. During training, we applied Q-learning updates to samples of experience drawn at random from the pool of stored samples. After updating, Q-net performs an $\epsilon$ -greedy policy that allows it to select a random action with probability $\epsilon=0.1$. More details regarding the DQN algorithm can be found in~\cite{mnih2013playing}.
		
		\textbf{Q-net}:
		The input to deep Q-net is shape-indexed patches extracted based on ground-truth landmarks, which can provide Q-net with a priori the location of the occlusion on the human face. The parameters of deep Q-net are initialized by the CNNs of the pre-trained SRN, which avoids collecting too many error experiences in the initial phase of training and accelerates model convergence. In our tests, the prediction accuracies of both random initialization and initialization with SRN are essentially the same, and the initializations differ only in training speed.
		
		\textbf{Tracker}:
		SRN can be easily extended to video data by re-tasking the RNN module to memorize the temporal information across adjacent frames; thus, the equation~\ref{RNN} can be rewritten as
		\begin{eqnarray} \label{RNN-temporal}
		h^{t}=\mathrm{tanh}((\mathbf{x} \oplus h^{t-1})*W_{2}+b_{2}),
		\end{eqnarray}
		where $h^{t}$ represents the hidden state of the $t$-th frame and $\mathbf{x}$ is the feature map set extracted by the three convolutional layers. $W_{2}$ denotes the weight matrix of the input-to-hidden fully connected layer, $b_{2}$ is the bias and $\oplus$ is the concatenation operation. With this simple re-tasking, SRN can more consistently localize the landmarks of dynamic faces without requiring network structure changes.

\section{Experiments}
\subsection{Implementation Details} 
\begin{figure}[H]
	\begin{center}
		
		\includegraphics[width=1\linewidth]{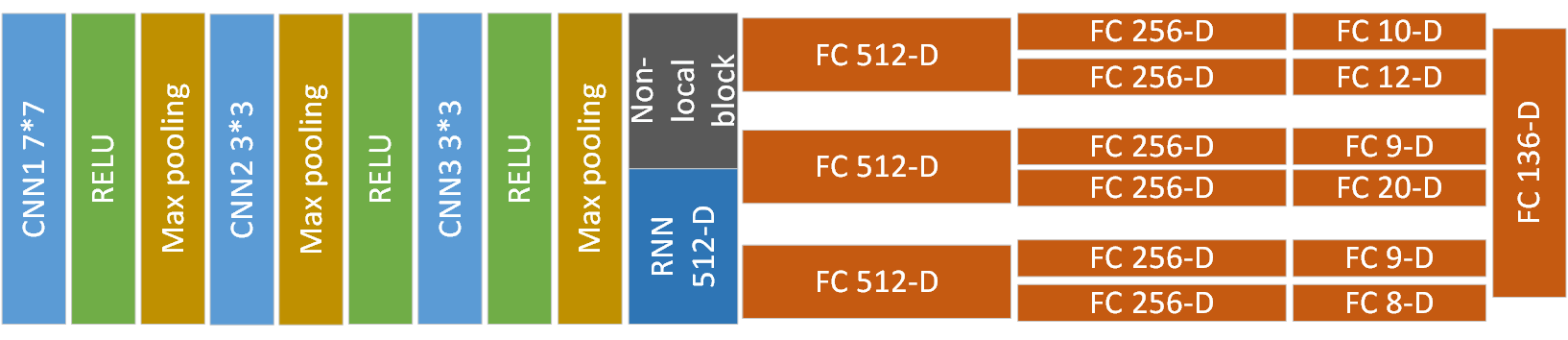}
	\end{center}
	\caption{Main components and parameters of the proposed SRN.}
	\label{NS}
\end{figure}

Figure \ref{NS} shows the structure of SRN for the prediction of $68$ landmarks. This diagram clearly shows the overall structure of the network and its detailed parameters. In addition, our open-source code will be updated and maintained soon, and additional details can be found in our code.

The SRN starts from a $7*7$ convolutional layer with a stride of $1$ and $32$ channels, followed by a max-pooling layer. Each of the two subsequent $3*3$ convolution layers, with a stride of $1$ and $64$ channels, are followed by a max-pooling layer behind each layer. In HSRM, the non-local block follows~\cite{Wang_2018_CVPR}. Each branch network of $g$ and $f$ is a fully connected layer with $512$-D and $256$-D, respectively. The dimensionality of each branch network in $d$ is the same as the number of landmarks in the corresponding facial component. The dimensionality of $\mu$ is the same as the total number of landmarks. Our SRN accepts $68$ raw patches according to the 300W annotation and each patch is cropped to a size of $36 \times 36$. For landmark tracking, synthesized faces are applied to fine-tune the pre-trained SRN. 

\subsection{Data Introduction and Evaluation Metrics} 

\textbf{Evaluation Datasets.} We evaluated the proposed SRN on six datasets: 300W~\cite{300W}, 300VW~\cite{DBLP:conf/iccvw/ShenZCKTP15}, COFW29~\cite{RCPR}, COFW68~\cite{ghiasi2015occlusion}, WFLW~\cite{LAB} and Masked 300W (containing masked faces).

\textit{300W}~\cite{300W}: Following the widely used evaluation setting \cite{MDM,SBR,LAB}, we integrate the AFW set ($337$ images), the LFPW training set ($2,000$ images) and the HELEN training set ($811$ images) to train the proposed SRN. We use the test sets of LFPW and HELEN as the Common set ($554$ images), the IBUG set as the Challenging set ($135$ images), and their as the Full set ($689$ images).

\textit{COFW29~\cite{RCPR} and COFW68}~\cite{ghiasi2015occlusion}:
The Caltech Occluded Face in the Wild (COFW29) dataset~\cite{RCPR} consists of $1345$ training face images and $507$ test face images collected from the Internet, all of which are annotated with $29$ landmarks. Because only a labeled test set exists COFW68, we performed a cross-evaluation on this test set, whose faces re-annotated with $68$ landmarks by~\cite{ghiasi2015occlusion}. Note that we used 300W to train the SRN for the cross-evaluation.

\textit{WFLW~\cite{LAB}}:
Wider Facial Landmarks in-the-wild (WFLW) which contains $10,000$ images, introduces large pose, expression, and occlusion variance. Each image is annotated with $98$ landmarks and $6$ attributes. We tested the SRN on a subset of WFLW occlusions to evaluate the robustness of the SRN against heavy occlusions.

\textit{Masked 300W}: Although COFW68 contains various occluded faces, it lacks heavily occluded faces, such as faces wearing a mask. Therefore, we simulated masked face images to synthesize the Masked 300W dataset by following SAAT~\cite{SAAT}. Note that the SRN was trained using 300W dataset without using any training images from the Masked 300W dataset.

\textit{300VW}~\cite{DBLP:conf/iccvw/ShenZCKTP15}: Typically used for facial landmark tracking, and contains $114$ videos (50 training videos and 64 test videos) captured under various conditions. Following the methods \cite{DBLP:journals/pami/LiuLFZ18,FHR,zhu2020towards,STKI}, we divided the $64$ test videos into three subsets: Category 1, Category 2 and Category 3. Then following these methods, 300W was applied to pre-train our SRN. Moreover, we used the same method as Masked 300W to generate masked-face videos to evaluate the effect of data synthesis on the SRN. Following the above methods, we used indices provided by 300VW organizers~\cite{DBLP:conf/iccvw/ShenZCKTP15} to remove corresponding frames from the evaluation set to obtain a fair comparison.

\textbf{Evaluation Metrics.}
We used the {Normalized Mean Error (NME)} and cumulative error distribution (CED) curves as evaluation metrics in our experiments, where the average point-to-point Euclidean error normalized by the inter-ocular distance and the inter-pupil distance were used as the error measure. Because different metrics have be used in some state-of-the-art methods~\cite{VALLE2019102846, kumar2020luvli}, we also used other metrics that are consistent with these methods~\cite{VALLE2019102846, kumar2020luvli}, such as inter-pupil normalization and bounding box normalization. The averaged NMEs of each test set determined the final performance. 
\subsection{Comparison with Existing Approaches}
\textbf{Evaluation on occlusion data.}
To evaluate the robustness of our SRN against occlusion, we compared the top performance methods~{\cite{VALLE2019102846,kumar2020luvli,huang2020propagationnet,valle2020multi}} on occlusion datasets COFW29, COFW68 and WFLW, where COFW68 was used to conduct cross-data experiments. For COFW29, we used the inter-pupil distance consistent with 3DDE~\cite{VALLE2019102846} to normalize the prediction error, and {we reported recall percentage at $80$ percent precision by following~\cite{valle2020multi}}. For COFW68, we used geometric mean of the width and height of the ground-truth bounding box, which is consistent with LUVLi~\cite{kumar2020luvli}. For the WFLW occlusion subset, we followed the LAB~\cite{LAB} and used the inter-ocular normalization. Table~\ref{R22} demonstrates that the proposed SRN achieves the state-of-the-art performance on occluded data. The only method that outperformed ours (PropNet~\cite{huang2020propagationnet}) requires a greater number of parameters.
 \begin{table} [H]
 	\begin{center}
 		\caption{Comparison to state-of-the-art methods on COFW29, COFW68, and WFLW. Note that WFLW$^{*}$ denotes the occlusion subset of WFLW.}
 		\resizebox{13cm}{!}{
 			\begin{tabular}{  c    c  c  c c c }
 				\toprule
 				{Datasets} &\multicolumn{2}{c}{{COFW29}}& { COFW68}  & WFLW$^{*}$ & WFLW \\
 				\hline 
 				Metrics & NME$_{pupil}$ & Recall & NME$_{bbx}$ & NME$_{ocular}$ &NME$_{ocular}$ \\ 
 				\hline
 				3DDE~\cite{VALLE2019102846} & 5.11 & 63.89 & - & 5.77 & 4.68 \\ 
 				MNN ~\cite{valle2020multi}  & 5.04 & 72.12& - & - & - \\
 				LUVLi~\cite{kumar2020luvli}  & - &  - & 2.75 & 5.29 & 4.37 \\ 
 				PropNet~\cite{huang2020propagationnet} & 3.71 &-  & - & 4.58 & 4.05 \\
 				\hline
 				SRN &   4.85 & 61.25 & 2.11 & 5.31 & 4.86\\
 				\bottomrule
 			\end{tabular}
 		}\label{R22}
 	\end{center}
 \end{table}

\begin{table} [t]
	\begin{center}
		\caption{Comparison to state-of-the-arts on Masked 300W~(68 landmarks). Note that  Masked 300W is only used for cross-dataset evaluation, not training.
		}
		\label{C300W}
		\resizebox{8.5cm}{!}{
			\begin{tabular}{  c   c  c  c  }
				\toprule
				\textbf{Methods} & \textbf{Challenging} &\textbf{ Common }& \textbf{ Full}   \\
				\hline
				CFSS~\cite{CFSS}  & 19.98 & 11.73 & 13.35 \\
				SBR~\cite{FHR}   & 13.28 & 8.72 & 9.6\\
				MDM~\cite{MDM}& 11.67 & 7.66 & 8.44 \\ 
				SHG~\cite{SHG} & 13.52 & 8.17 & 9.22 \\
				DHGN~\cite{DHGN} & 12.19 & 8.98 & 9.61 \\
				FHR~\cite{FHR}  & 11.28 & 7.02 & 7.85\\
				LAB~\cite{LAB} & 9.59 & 6.07 & 6.76 \\
				\hline
				{SRN} & \textbf{9.28} & \textbf{5.78} & \textbf{6.46} \\
				\bottomrule
			\end{tabular}
		}
	\end{center}
\end{table}
\begin{figure*}[b]
	\begin{center}
		\includegraphics[width=0.95\linewidth]{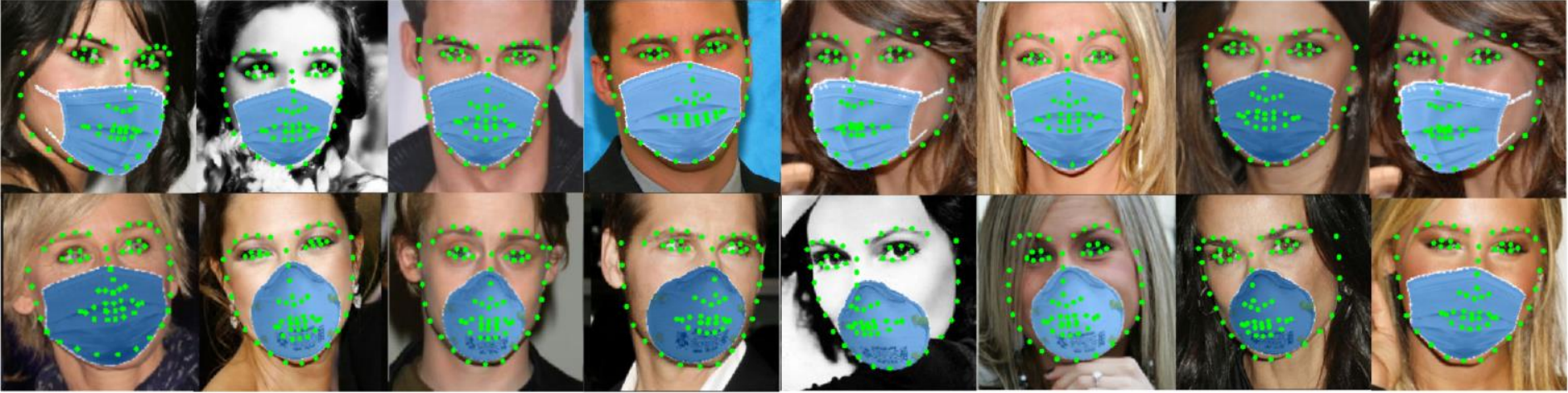}
	\end{center}
	\caption{
		Qualitative prediction results on Masked 300W dataset. 
	}
	\label{Masked_face}
\end{figure*}
\textbf{Evaluation on Masked 300W.}
To further highlight the robustness of SRN with respect to masked faces, we trained our model using 300W~\cite{300W} and evaluated its performance on masked faces. To this end, we generated an equivalent Masked 300W dataset, which was produced by placing a mask on each 300W face. The prediction results are presented in Figure \ref{Masked_face}. Even under severe occlusion, the proposed method still infers reasonable face shapes. These results indicate that the proposed SRN can efficiently handle masked faces.

Table~\ref{C300W} shows the comparison results with existing methods. Note that  Masked 300W is only used for cross-dataset evaluation in the test phase, not during the training stage. Compared with LAB~\cite{LAB} that considers face boundary information, our method displays decreases in error of 3.23$\%$, 4.78$\%$ and 4.43$\%$ on the Masked 300W Challenging, Common, and Full sets, respectively. This occurs because face boundary information is still affected by occlusion and partial observability. In contrast, SRN reasons the structural relations that can efficiently constrain the structure of facial shapes to overcome various occlusions.

\textbf{Evaluations on COFW68.}
Table \ref{cofw} compares the averaged errors and failure rates of our SRN to other existing methods on COFW68, which contains more occluded landmarks. The results indicate that the SRN outperforms most state-of-the-art methods by a large margin. Even without the use of occluded face augmentation, our SRN still achieves the best performance. Its lowest failure rate also proves that our method achieves outstanding generalization performance on occluded faces. 
\begin{table} [h]
	\begin{center}
		\caption{Comparison of averaged errors and failure
			rates (threshold at 0.08\%) for the COFW68 dataset. Our SRN achieves occlusion-robust landmark localization. Note that SRN (w/o OFA) denotes the proposed SRN without using occluded face augmentation. 
		}
		\label{cofw}
		\resizebox{9cm}{!}{
			\begin{tabular}{  c   c  c  c  }
				\toprule
				\textbf{Method} & \textbf{Averaged Error} &\textbf{ Failure Rate ($\%$) }  \\
				\hline
				SDM~\cite{SDM} & 8.77 & 24.32  \\
				RPP ~\cite{yang2015robust}  & 7.52 & 16.20  \\
				CFAN~\cite{zhang2014coarse} & 8.38 & 19.14   \\
				TCDCN~\cite{zhang2015learning} & 8.05 & 15.31 \\
				MDM~\cite{MDM} & 6.32 & 10.31  \\
				ODN~\cite{ODN} & 5.87 & 9.43  \\
				DHGN~\cite{DHGN}& 5.29 & 6.94 \\
				\hline			
				SRN (w/o OFA)  & 4.93 & 4.71  \\
				SRN  & \textbf{4.61} & \textbf{4.34}  \\
				\bottomrule
			\end{tabular}
		}
	\end{center}
\end{table}


\textbf{Evaluation on 300W.}
 \begin{table} [tp]
	\begin{center}
		\caption{Comparison to state-of-the-art methods on 300W. {Note that SRN+HG uses the initial landmarks provided by a one-stage hourglass network.}
		}
		\resizebox{12cm}{!}{
			\begin{tabular}{  c   c  c  c c }
				\toprule
				\textbf{Methods} & \textbf{Challenging} &\textbf{ Common }& \textbf{ Full}  & \textbf{ heatmap prediction} \\
				\hline
				\multicolumn{5}{c}{\textbf{ Inter-ocular}}\\
				\hline
				MDM~\cite{MDM} & 8.87 & 3.74 & 4.78 &  no \\ 
				HGs~\cite{SHG} & 7.52 & 3.17 & 4.01&  yes \\
				
				RDN~\cite{RDN}  & 7.04 & 3.31 & 4.23   & no\\
				SBR~\cite{SBR} & 7.58 & 3.28 & 4.10 &  yes\\
				ODN~\cite{ODN}  & 6.67 & 3.56 & 4.17& no \\
				DHGN~\cite{DHGN} & 6.23 & 3.38 & 3.95& yes \\
				3FabRec \cite{Browatzki_2020_CVPR} & 5.74 & 3.36 & 3.82 &yes \\
				Chandran \textit{et.al} \cite{chandran2020attention} & 7.04 & 2.83 & 4.23 &yes\\
				LUVLi~\cite{kumar2020luvli} & 5.16 & 2.76 & 3.23  &yes\\
				PropagationNet \cite{huang2020propagationnet} &3.99  & 2.67 & 2.93 &yes\\
				\hline
				{SRN} (ours) & {5.86} & 3.08 & 3.62  &no\\
				{SRN+HG} (ours) & {5.38} & 3.03 & 3.49  &yes\\
				\hline
				\hline
				\multicolumn{5}{c}{\textbf{ Inter-pupil}}\\
				\hline
				CFSS~\cite{ZhuLLT15}   & 9.98 & 4.73 & 5.76 & no\\
				3DDE~\cite{VALLE2019102846}  & 7.10 & 3.73 & 4.39 &yes\\
				AWing~\cite{wang2019adaptive}  & 6.52 & 3.77 & 4.31 & yes\\
				PropagationNet \cite{huang2020propagationnet} & 5.75 &  3.70& 4.10  &yes\\
				\hline
				{SRN} (ours) & {7.69} & 3.97 & 4.70  &no\\
				{SRN+HG} (ours) & {7.04} & 3.78 & 4.42  &yes\\
				\bottomrule
			\end{tabular}
		}\label{T300W}
	\end{center}
\end{table}
We compare the proposed SRN with various state-of-the-arts shown in Table~\ref{T300W}. {There are four state-of-the-art methods (LUVLi~\cite{kumar2020luvli}, PropagationNet~\cite{huang2020propagationnet}, 3DDE~\cite{VALLE2019102846} and Awing~\cite{wang2019adaptive}) that are significantly stronger than the proposed SRN on the challenge set. Their performance gains in 300W are significant.} However, we found a very interesting fact — those methods that outperform ours are all heatmap-based models. Heatmap-based methods model landmark localization as a classification task based on pixels, which are more robust to large poses than shape regression. Most of the hard samples in this Challenge set involved large pose rather than occlusion. We further found that {most} of these heatmap-based methods unsurprisingly use stacked deep networks, such as stacked U-net~\cite{ronneberger2015u}, or a stacked hourglass (SHG) network~\cite{SHG}. Therefore, these methods face two disadvantages: 1) their large computational overhead limits their usage scenarios and 2) heatmap prediction does not involve shape constraint, making it difficult to for these methods overcome the occlusion problem.  
In particular, the best method~\cite{huang2020propagationnet} that outperforms ours requires a greater number of parameters ($36.3$M) than ours ($19.89$M), which may explain why it achieves better performance than SRN.

{To further improve the performance of the SRN on large-pose faces, we follow 3DDE using a simple heatmap-based model (\textit{i.e.} one-stage hourglass network (HG)~\cite{SHG}) to provide the initial landmarks (\textit{i.e.} SRN+HG). We can see that the performance of the SRN is further improved when it uses the initial landmarks providing by the hourglass network. Since addressing the large pose problem is beyond the scope of this paper, we only propose a solution and do not discuss and compare more.}

\textbf{Evaluation on video dataset 300VW.}
\begin{table}[t]
	\centering
	\caption{ Comparison of averaged errors between the proposed method and state-of-the-art methods on the 300VW~(68-landmarks) video dataset, where HOFS denotes "hard occluded face synthesis".}
	\label{T300VW}
	\begin{tabular}{cccc}  
		\toprule
		\textbf{Methods} & \textbf{Category 1} & \textbf{Category 2} & \textbf{Category 3} \\
		\midrule
		iCCR \cite{DBLP:conf/eccv/Sanchez-LozanoM16}  & {6.71 }& {4.00} & {12.75}  \\
		SBR \cite{SBR} & 5.09 & 4.76 & 10.21\\
		TSTN  \cite{DBLP:journals/pami/LiuLFZ18}  &{{5.36}} & {{{4.51}}} & {{12.84}}  \\
		STRRN \cite{zhu2020towards} & 5.03 & 4.74 & 6.63\\
		STKI \cite{STKI}  & 5.04 & 4.57 & 6.11\\
		FHR+STA\cite{FHR}   & 4.42 & 4.18 & 5.98\\
		Yin \textit{et.al.} \cite{yin2020exploiting}   & 3.34 & 3.51 & 4.33\\
		FAB  \cite{sun2019fab}   & 4.24 & 5.67 & 3.16\\
		\hline
		{SRN+HOFS}     &  {3.87}  &  {3.96}  &  {5.15}\\
		
		\bottomrule
	\end{tabular}
\end{table}
Table~\ref{T300VW} lists the averaged NMEs of state-of-art methods for the three categories. In all categories, only Yin \textit{et.al.} \cite{yin2020exploiting} outperforms the SRN. It should be noted that the baseline \cite{yin2019capturing} used in method Yin \textit{et.al.} \cite{yin2020exploiting} has multiple consecutive frames for each prediction, and the features of each frame are extracted by a four-stage stacked HG network and then fed into the sub-networks, which makes the model very inefficient. Although the authors did not report the inference speed or provide open source code, we found that four-stage stacked hourglass networks ($21$ FPS) run much slower than the SRN ($31$ FPS). Additionally, each prediction of Yin \textit{et.al.} \cite{yin2020exploiting} requires feature extraction for multiple frames at the same time, resulting in higher computational cost. Therefor, the SRN is still competitive.

\begin{figure*}[t]
	
	\begin{center}
		
		\includegraphics[width=1\linewidth]{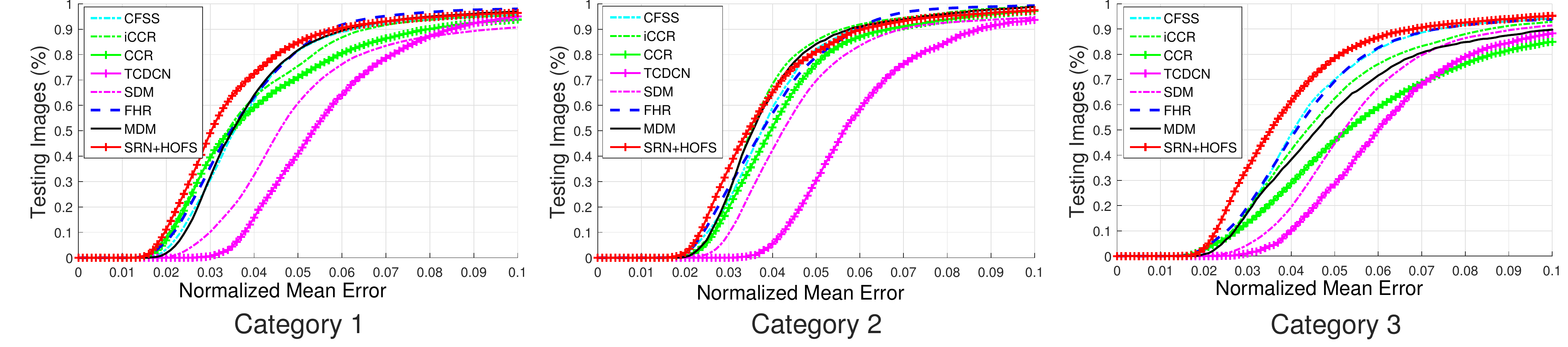}
	\end{center}
	\vspace{-0.7cm}
	\caption{CED curves of SRN+HOFS for all three categories in 300VW. }
	\label{fig:CED}
\end{figure*}

\begin{figure}[t]
	
	\begin{center}
		
		\includegraphics[width=1\linewidth]{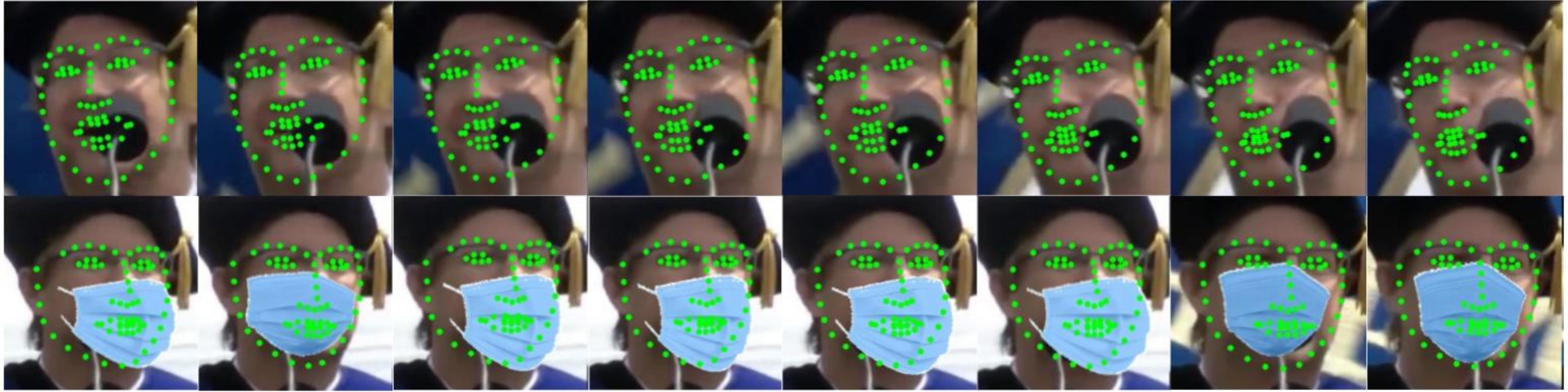}
	\end{center}
	\caption{Tracking results on 300VW. The first row denotes the results of tracking dynamic occlusions, and the second row represents the results of tracking masked faces. The proposed SRN exhibits high-quality facial structures under these two types of typical occlusions.}
	\label{300VW_images}
\end{figure}

Figure~\ref{fig:CED} shows the CED curves comparing our method compared to state-of-the-art methods~\cite{SDM,CFSS,FHR,MDM,ZhangPAMI15,DBLP:conf/eccv/Sanchez-LozanoM16}. From these results, it can be concluded that SRN significantly outperforms most facial landmark localization methods by a large margin. 

Figure \ref{300VW_images} illustrates some qualitative tracking results. These results indicate that our SRN captures the structural relations to constrain a facial shape, which improves its robustness to the occlusion of dynamic faces.

It should be noted that to prove the extendibility of SRN, all of the above experiments were performed by changing only the input of the RNN of the image-based model as formalized in Equation~\ref{RNN-temporal}; the initial face shape and the maximum number of iterations remain unchanged. In the computational efficiency section below, we evaluate the impact of different initial shapes and maximum iteration steps on model performance and running speed.

\textbf{Computational Efficiency.}
Table~\ref{tab:FPS} tabulates inference speed comparisons and indicates that our SRN can run at $31.53$ frames per second on a single 1080Ti GPU, which outperforms most heatmap regression models. Note that the $31.53$ FPS is achieved in the video-based mode using a mean shape and $3$ iterations and excludes the data loading and face detection steps. 

\begin{table}[H]
	\centering
	\caption{Comparison involving inference speed. FPS denotes the number of images processed per second. }
	\label{tab:FPS}
	\begin{tabular}{ccccc}  
		\toprule
		Methods & LAB\cite{LAB} & DHGN\cite{DHGN} & SBR\cite{SBR} & SRN \\
		\midrule
		FPS  & 12.46 & 13.32 & 19.82 & 31.53 \\
		\bottomrule
	\end{tabular}
\end{table}
%

\subsection{Ablation Study}

%
%
%

\textbf{Hierarchical Structural Relation Module} 
To further investigate the ablation experiments regarding structural relations, we evaluate each part of the proposed hierarchical structural relation module (HSRM) on Masked 300W. To this end, $g$, $f$, $d$, and $\mu$ are removed from the HSRM for evaluation in sequence, as shown in Table \ref{tab:HSRM}. Taking $g$ as an example, SRN w/o $g$ means that the proposed SRN is implemented without equipping the layers in $g$. Table \ref{tab:HSRM} demonstrates that the hierarchical structural relation from coarse to fine is the most effective.

\begin{table}[t]
	\centering
	\caption{Ablation experiments involving the proposed hierarchical structural relation module (HSRM) on Masked 300W. Here, w/o denotes the implementation of our SRN without equipping the corresponding layers.}
	\label{tab:HSRM}
	\begin{tabular}{cccc}  
		\toprule
		{Methods} & {Challenging} &{ Common }& { Full}   \\
		\hline
		SRN w/o $g$   & 11.72 & 6.46 & 7.49 \\
		SRN w/o $f$   & 10.20 & 6.13 & 6.92 \\
		SRN w/o $d$   & 12.02 & 6.86 & 7.86 \\
		SRN w/o $\mu$   & 9.68 & 5.96 & 6.69 \\
		SRN   & 9.28 & 5.78 & 6.46\\
		
		\bottomrule
	\end{tabular}
	
\end{table}

\begin{table}[H]
	\centering
	\caption{Averaged NMEs on Masked 300W for evaluation of the non-local block. Here SRN (w/o NLB) and SRN (w NLB) denote the trained SRN with and without the non-local block.}
	\label{tab:Non-local}
	\begin{tabular}{cccc}  
		\toprule
		{Methods} & {Challenging} &{ Common }& { Full}   \\
		\hline
		SRN (w/o NLB)  & 9.47 & 5.81 & 6.52 \\
		SRN (w NLB)  & 9.28 & 5.78 & 6.46\\
		\bottomrule
	\end{tabular}
	
\end{table}

\textbf{Non-local Operation.}
A non-local operation is introduced to explain the global relations in our network. We verify the potential of the non-local operation that captures long- and short-distance dependencies well. In Table \ref{tab:Non-local}, {we can see that the performance gains of the non-local operation, on three test sets (Challenging, Common, Full) of Masked 300W, achieves $0.19$, $0.04$ and $0.06$ in NME, respectively.}

\textbf{Hard occluded face synthesis} 
To evaluate the impact of occluded face synthesis, Table \ref{tab:HOFS} presents the localization errors of our SRN when using different forms of occluded face synthesis on 300VW, and demonstrates that synthesizing hard occluded faces can efficiently improve robustness to occlusion. This is because introducing occluded faces diversifies the training data.

	\begin{table}[H]
	\centering
	\caption{Ablation experiments involving occluded face augmentation on 300VW, where RSF and HOFS denote randomly synthesizing occlusion and hard occluded face synthesis.}
	\label{tab:HOFS}
	\begin{tabular}{cccc}  
		\toprule
			\multicolumn{4}{c}{\textbf{ 300VW test set }}\\
				Methods & Category 1 & Category 2 & Category 3 \\
				\midrule
				SRN w/o OFA     &  {4.01}  &  {4.10}  &  {5.52}\\
				SRN + RSF & 3.92 & 4.02 & 5.16 \\
				SRN + HOFS   & 3.87 & 3.96 & 5.15 \\

				\hline
				\multicolumn{4}{c}{\textbf{ 300VW test set (Masked videos) }}\\
				\hline
				SRN w/o DA     &  {7.31}  &  {7.62}  &  {8.52}\\
				SRN + RSF & 6.82 & 6.73 & 8.09 \\
				SRN + HOFS   & 6.03 & 5.83 & 7.35 \\
		
		\bottomrule
	\end{tabular}
\end{table}

\section{Conclusion}

We have proposed a structural relation network (SRN) to provide robust occlusion facial landmark localization. Our architecture significantly improves robustness to occlusion and partial observability in facial images. 
The advantages of the proposed method are as follows: (1) SRN is lightweight and can be easily extended; (2) SRN imposes facial shape constraints into the network structure and does not require additional sub-networks or sub-tasks; and (3) we use reinforcement learning to augment 'hard' occlusion samples in videos, which addresses the lack of occlusion clips present in existing landmark tracking video datasets.  
		
SRN has some shortcomings, mainly regarding the lack of fine-grained fitting for faces having extreme large poses. The large poses induce the cascade regression algorithm to easily fall into local optima.

Future work includes implementing robust tracking that can handle low-quality video, containing for example, low resolution and/or drastic light changes.

\section{Acknowledgements}
This work is supported in part by Shanghai science and technology committee under grant No.19511121002. We appreciate the High Performance Computing Center of Shanghai University, and Shanghai Engineering Research Center of Intelligent Computing System No.19DZ2252600 for providing the computing resources and technical support.

\bibliographystyle{elsarticle-num} 
\bibliography{PR}

\begin{thebibliography}{10}
\expandafter\ifx\csname url\endcsname\relax
  \def\url#1{\texttt{#1}}\fi
\expandafter\ifx\csname urlprefix\endcsname\relax\def\urlprefix{URL }\fi
\expandafter\ifx\csname href\endcsname\relax
  \def\href#1#2{#2} \def\path#1{#1}\fi

\bibitem{jiang20183d}
L.~Jiang, J.~Zhang, B.~Deng, H.~Li, L.~Liu, 3d face reconstruction with
  geometry details from a single image, TIP 27~(10) (2018) 4756--4770.

\bibitem{gecer2019ganfit}
B.~Gecer, S.~Ploumpis, I.~Kotsia, S.~Zafeiriou, Ganfit: Generative adversarial
  network fitting for high fidelity 3d face reconstruction, in: CVPR, 2019, pp.
  1155--1164.

\bibitem{PRNET}
Y.~Feng, F.~Wu, X.~Shao, Y.~Wang, X.~Zhou, Joint 3d face reconstruction and
  dense alignment with position map regression network, arXiv:1803.07835
  (2018).

\bibitem{zhao2018towards}
J.~Zhao, Y.~Cheng, Y.~Xu, L.~Xiong, J.~Li, F.~Zhao, K.~Jayashree, S.~Pranata,
  S.~Shen, J.~Xing, et~al., Towards pose invariant face recognition in the
  wild, in: CVPR, 2018, pp. 2207--2216.

\bibitem{kang2018pairwise}
B.-N. Kang, Y.~Kim, D.~Kim, Pairwise relational networks for face recognition,
  in: ECCV, 2018, pp. 628--645.

\bibitem{song2019geometry}
L.~Song, J.~Cao, L.~Song, Y.~Hu, R.~He, Geometry-aware face completion and
  editing, in: AAAI, Vol.~33, 2019, pp. 2506--2513.

\bibitem{wang2019example}
M.~Wang, G.-Y. Yang, R.~Li, R.-Z. Liang, S.-H. Zhang, P.~M. Hall, S.-M. Hu,
  Example-guided style-consistent image synthesis from semantic labeling, in:
  CVPR, 2019, pp. 1495--1504.

\bibitem{MDM}
G.~Trigeorgis, P.~Snape, M.~A. Nicolaou, E.~Antonakos, S.~Zafeiriou, Mnemonic
  descent method: A recurrent process applied for end-to-end face alignment,
  in: CVPR, 2016, pp. 4177--4187.

\bibitem{SBR}
X.~Dong, S.-I. Yu, X.~Weng, S.-E. Wei, Y.~Yang, Y.~Sheikh,
  Supervision-by-registration: An unsupervised approach to improve the
  precision of facial landmark detectors, in: CVPR, 2018, pp. 360--368.

\bibitem{3DDFA}
S.~HL, et~al., Face alignment across large poses: A 3d solution, in: CVPR,
  2016, pp. 146--155.

\bibitem{FAN}
A.~Bulat, G.~Tzimiropoulos, How far are we from solving the 2d \& 3d face
  alignment problem?(and a dataset of 230,000 3d facial landmarks), in: ICCV,
  2017, pp. 1021--1030.

\bibitem{LAB}
W.~Wu, C.~Qian, S.~Yang, Q.~Wang, Y.~Cai, Q.~Zhou, Look at boundary: A
  boundary-aware face alignment algorithm, in: CVPR, 2018, pp. 2129--2138.

\bibitem{SDM}
X.~Xiong, F.~De~la Torre, Supervised descent method and its applications to
  face alignment, in: CVPR, 2013, pp. 532--539.

\bibitem{LBF}
S.~Ren, X.~Cao, Y.~Wei, J.~Sun, Face alignment at 3000 fps via regressing local
  binary features, in: CVPR, 2014, pp. 1685--1692.

\bibitem{CFSS}
S.~Zhu, C.~Li, C.~Change~Loy, X.~Tang, Face alignment by coarse-to-fine shape
  searching, in: CVPR, 2015, pp. 4998--5006.

\bibitem{santoro2017simple}
A.~Santoro, D.~Raposo, D.~G. Barrett, M.~Malinowski, R.~Pascanu, P.~Battaglia,
  T.~Lillicrap, A simple neural network module for relational reasoning, in:
  NeurIPS, 2017, pp. 4967--4976.

\bibitem{liu2019semantic}
Z.~Liu, X.~Zhu, G.~Hu, H.~Guo, M.~Tang, Z.~Lei, N.~M. Robertson, J.~Wang,
  Semantic alignment: Finding semantically consistent ground-truth for facial
  landmark detection, in: CVPR, 2019, pp. 3467--3476.

\bibitem{kumar2018disentangling}
A.~Kumar, R.~Chellappa, Disentangling 3d pose in a dendritic cnn for
  unconstrained 2d face alignment, in: CVPR, 2018, pp. 430--439.

\bibitem{RCPR}
X.~P. Burgos-Artizzu, P.~Perona, P.~Dollar, Robust face landmark estimation
  under occlusion, in: ICCV, 2013.

\bibitem{mnih2013playing}
V.~Mnih, K.~Kavukcuoglu, D.~Silver, A.~Graves, I.~Antonoglou, D.~Wierstra,
  M.~Riedmiller, Playing atari with deep reinforcement learning, arXiv preprint
  arXiv:1312.5602 (2013).

\bibitem{CootsAAM}
T.~F. Cootes, G.~J. Edwards, C.~J. Taylor, Active appearance models, TPAMI
  23~(6) (2001) 681--685.

\bibitem{ESR}
X.~Cao, Y.~Wei, F.~Wen, J.~Sun, Face alignment by explicit shape regression,
  IJCV 107~(2) (2014) 177--190.

\bibitem{PCR}
G.~Tzimiropoulos, Project-out cascaded regression with an application to face
  alignment, in: CVPR, 2015, pp. 3659--3667.

\bibitem{zhang2014coarse}
J.~Zhang, S.~Shan, M.~Kan, X.~Chen, Coarse-to-fine auto-encoder networks (cfan)
  for real-time face alignment, in: ECCV, Springer, 2014, pp. 1--16.

\bibitem{HongMPT13}
Z.~Hong, X.~Mei, D.~Prokhorov, D.~Tao, Tracking via robust multi-task
  multi-view joint sparse representation, in: ICCV, 2013, pp. 649--656.

\bibitem{TSR}
J.~Lv, X.~Shao, J.~Xing, C.~Cheng, X.~Zhou, A deep regression architecture with
  two-stage re-initialization for high performance facial landmark detection,
  in: CVPR, 2017, pp. 3317--3326.

\bibitem{3DMM}
V.~Blanz, T.~Vetter, Face recognition based on fitting a 3d morphable model,
  TPAMI 25~(9) (2003) 1063--1074.

\bibitem{PIFA}
A.~Jourabloo, X.~Liu, Pose-invariant face alignment via cnn-based dense 3d
  model fitting, IJCV 124~(2) (2017) 187--203.

\bibitem{JOINT}
F.~Liu, D.~Zeng, Q.~Zhao, X.~Liu, Joint face alignment and 3d face
  reconstruction, in: ECCV, Springer, 2016, pp. 545--560.

\bibitem{VALLE2019102846}
R.~Valle, J.~M. Buenaposada, A.~Valdés, L.~Baumela, Face alignment using a 3d
  deeply-initialized ensemble of regression trees, CVIU 189 (2019) 102846.

\bibitem{kumar2020luvli}
A.~Kumar, T.~K. Marks, W.~Mou, Y.~Wang, M.~Jones, A.~Cherian, T.~Koike-Akino,
  X.~Liu, C.~Feng, Luvli face alignment: Estimating landmarks' location,
  uncertainty, and visibility likelihood, in: CVPR, 2020, pp. 8236--8246.

\bibitem{huang2020propagationnet}
X.~Huang, W.~Deng, H.~Shen, X.~Zhang, J.~Ye, Propagationnet: Propagate points
  to curve to learn structure information, in: CVPR, 2020, pp. 7265--7274.

\bibitem{DBLP:conf/eccv/Sanchez-LozanoM16}
E.~S{\'{a}}nchez{-}Lozano, B.~Mart{\'{\i}}nez, G.~Tzimiropoulos, M.~F. Valstar,
  Cascaded continuous regression for real-time incremental face tracking, in:
  ECCV, 2016, pp. 645--661.

\bibitem{DBLP:conf/eccv/GuoLZ18}
M.~Guo, J.~Lu, J.~Zhou, Dual-agent deep reinforcement learning for deformable
  face tracking, in: ECCV, 2018, pp. 783--799.

\bibitem{DBLP:journals/pami/LiuLFZ18}
H.~Liu, J.~Lu, J.~Feng, J.~Zhou, Two-stream transformer networks for
  video-based face alignment, TPAMI 40~(11) (2018) 2546--2554.

\bibitem{STKI}
C.~Zhu, X.~Li, J.~Li, G.~Ding, W.~Tong, Spatial-temporal knowledge integration:
  Robust self-supervised facial landmark tracking, in: ACM MM, 2020, p.
  4135–4143.

\bibitem{FHR}
Y.~Tai, Y.~Liang, X.~Liu, L.~Duan, J.~Li, C.~Wang, F.~Huang, Y.~Chen, Towards
  highly accurate and stable face alignment for high-resolution videos, in:
  AAAI, Vol.~33, 2019, pp. 8893--8900.

\bibitem{DBLP:conf/eccv/PengFWM16}
X.~Peng, R.~S. Feris, X.~Wang, D.~N. Metaxas, A recurrent encoder-decoder
  network for sequential face alignment, in: ECCV, 2016, pp. 38--56.

\bibitem{zhu2020towards}
C.~Zhu, H.~Liu, Z.~Yu, X.~Sun, Towards omni-supervised face alignment for large
  scale unlabeled videos., in: AAAI, 2020, pp. 13090--13097.

\bibitem{sun2019fab}
K.~Sun, W.~Wu, T.~Liu, S.~Yang, Q.~Wang, Q.~Zhou, Z.~Ye, C.~Qian, Fab: A robust
  facial landmark detection framework for motion-blurred videos, in: CVPR,
  2019, pp. 5462--5471.

\bibitem{yin2020exploiting}
S.~Yin, S.~Wang, X.~Chen, E.~Chen, Exploiting self-supervised and
  semi-supervised learning for facial landmark tracking with unlabeled data,
  in: ACM MM, 2020, pp. 2991--2998.

\bibitem{buades2005non}
A.~Buades, B.~Coll, J.-M. Morel, A non-local algorithm for image denoising, in:
  CVPR, Vol.~2, IEEE, 2005, pp. 60--65.

\bibitem{Wang_2018_CVPR}
X.~Wang, R.~Girshick, A.~Gupta, K.~He, Non-local neural networks, in: CVPR,
  2018.

\bibitem{mnih2015human}
V.~Mnih, K.~Kavukcuoglu, D.~Silver, A.~A. Rusu, J.~Veness, M.~G. Bellemare,
  A.~Graves, M.~Riedmiller, A.~K. Fidjeland, G.~Ostrovski, et~al., Human-level
  control through deep reinforcement learning, nature 518~(7540) (2015)
  529--533.

\bibitem{lin1992reinforcement}
L.-J. Lin, Reinforcement learning for robots using neural networks, Carnegie
  Mellon University, 1992.

\bibitem{300W}
C.~Sagonas, E.~Antonakos, G.~Tzimiropoulos, S.~Zafeiriou, M.~Pantic, 300 faces
  in-the-wild challenge: database and results, IVC 47 (2016) 3--18.

\bibitem{DBLP:conf/iccvw/ShenZCKTP15}
J.~Shen, S.~Zafeiriou, G.~G. Chrysos, J.~Kossaifi, G.~Tzimiropoulos, M.~Pantic,
  The first facial landmark tracking in-the-wild challenge: Benchmark and
  results, in: ICCVW, 2015, pp. 1003--1011.

\bibitem{ghiasi2015occlusion}
G.~Ghiasi, C.~C. Fowlkes, Occlusion coherence: Detecting and localizing
  occluded faces, arXiv preprint arXiv:1506.08347 (2015).

\bibitem{SAAT}
C.~Zhu, X.~Li, J.~Li, S.~Dai, Improving robustness of facial landmark detection
  by defending against adversarial attacks, in: ICCV, 2021, pp. 11751--11760.

\bibitem{valle2020multi}
R.~Valle, J.~M. Buenaposada, L.~Baumela, Multi-task head pose estimation
  in-the-wild, TPAMI (2020).

\bibitem{SHG}
A.~Newell, K.~Yang, J.~Deng, Stacked hourglass networks for human pose
  estimation, in: ECCV, Springer, 2016, pp. 483--499.

\bibitem{DHGN}
H.~Zhu, H.~Liu, C.~Zhu, Z.~Deng, X.~Sun, Learning spatial-temporal deformable
  networks for unconstrained face alignment and tracking in videos, Pattern
  Recognition 107 (2020) 107354.

\bibitem{yang2015robust}
H.~Yang, X.~He, X.~Jia, I.~Patras, Robust face alignment under occlusion via
  regional predictive power estimation, TIP 24~(8) (2015) 2393--2403.

\bibitem{zhang2015learning}
Z.~Zhang, P.~Luo, C.~C. Loy, X.~Tang, Learning deep representation for face
  alignment with auxiliary attributes, TPAMI 38~(5) (2015) 918--930.

\bibitem{ODN}
M.~Zhu, D.~Shi, M.~Zheng, M.~Sadiq, Robust facial landmark detection via
  occlusion-adaptive deep networks, in: CVPR, 2019, pp. 3486--3496.

\bibitem{RDN}
H.~Liu, J.~Lu, M.~Guo, S.~Wu, J.~Zhou, Learning reasoning-decision networks for
  robust face alignment, TPAMI (2018).

\bibitem{Browatzki_2020_CVPR}
B.~Browatzki, C.~Wallraven, 3fabrec: Fast few-shot face alignment by
  reconstruction, in: CVPR, 2020.

\bibitem{chandran2020attention}
P.~Chandran, D.~Bradley, M.~Gross, T.~Beeler, Attention-driven cropping for
  very high resolution facial landmark detection, in: CVPR, 2020, pp.
  5861--5870.

\bibitem{ZhuLLT15}
S.~Zhu, C.~Li, C.~C. Loy, X.~Tang, Face alignment by coarse-to-fine shape
  searching, in: CVPR, 2015, pp. 4998--5006.

\bibitem{wang2019adaptive}
X.~Wang, L.~Bo, L.~Fuxin, Adaptive wing loss for robust face alignment via
  heatmap regression, in: ICCV, 2019, pp. 6971--6981.

\bibitem{ronneberger2015u}
O.~Ronneberger, P.~Fischer, T.~Brox, U-net: Convolutional networks for
  biomedical image segmentation, in: International Conference on Medical image
  computing and computer-assisted intervention, Springer, 2015, pp. 234--241.

\bibitem{yin2019capturing}
S.~Yin, S.~Wang, G.~Peng, X.~Chen, B.~Pan, Capturing spatial and temporal
  patterns for facial landmark tracking through adversarial learning., in:
  IJCAI, 2019, pp. 1010--1017.

\bibitem{ZhangPAMI15}
Z.~Zhang, P.~Luo, C.~C. Loy, X.~Tang, Learning deep representation for face
  alignment with auxiliary attributes, TPAMI 38~(5) (2016) 918--930.

\end{thebibliography}

\end{document}